\documentclass{ecai}  

\usepackage{latexsym}
\usepackage{amssymb}
\usepackage{amsmath}
\usepackage{amsthm}
\usepackage{booktabs}
\usepackage{enumitem}
\usepackage{graphicx}
\usepackage{color}
\usepackage{multirow}
\usepackage[normal]{caption2}

\begin{document}

\begin{frontmatter}

\title{MultiCounter: Multiple Action Agnostic Repetition Counting in Untrimmed Videos}



\author[A]{\fnms{Yin}~\snm{Tang}}
\author[A]{\fnms{Wei}~\snm{Luo}}
\author[B]{\fnms{Jinrui}~\snm{Zhang}}
\author[A]{\fnms{Wei}~\snm{Huang}}
\author[C]{\fnms{Ruihai}~\snm{Jing}}
\author[A]{\fnms{Deyu}~\snm{Zhang}\thanks{Corresponding Author. Email: zdy876@csu.edu.cn.}}

\address[A]{Central South University}
\address[B]{Tsinghua University}
\address[C]{Shanghai Transsion CO., LTD}


\begin{abstract}
Multi-instance Repetitive Action Counting (MRAC) aims to estimate the number of repetitive actions performed by multiple instances in untrimmed videos, commonly found in human-centric domains like sports and exercise. In this paper, we propose MultiCounter, a fully end-to-end deep learning framework that enables simultaneous detection, tracking, and counting of repetitive actions of multiple human instances. Specifically, MultiCounter incorporates two novel modules: 1) mixed spatiotemporal interaction for efficient context correlation across consecutive frames, and 2) task-specific heads for accurate perception of periodic boundaries and generalization for action-agnostic human instances. We train MultiCounter on a synthetic dataset called MultiRep that is generated from annotated real-world videos. Experiments on the MultiRep dataset validate the fundamental challenge of MRAC tasks and showcase the superiority of our proposed model. Compared to ByteTrack+RepNet, a solution that combines an advanced tracker with a single repetition counter, MultiCounter substantially improves Period-mAP by 41.0\%, reduces AvgMAE by 58.6\%, and increases AvgOBO 1.48 times. This sets a new benchmark in the field of MRAC. Moreover, MultiCounter runs in real-time on a commodity GPU server and is insensitive to the number of human instances in a video.
\end{abstract}
\end{frontmatter}


\section{Introduction}

\begin{figure}[h]
\centering
\includegraphics[width=8.5cm]{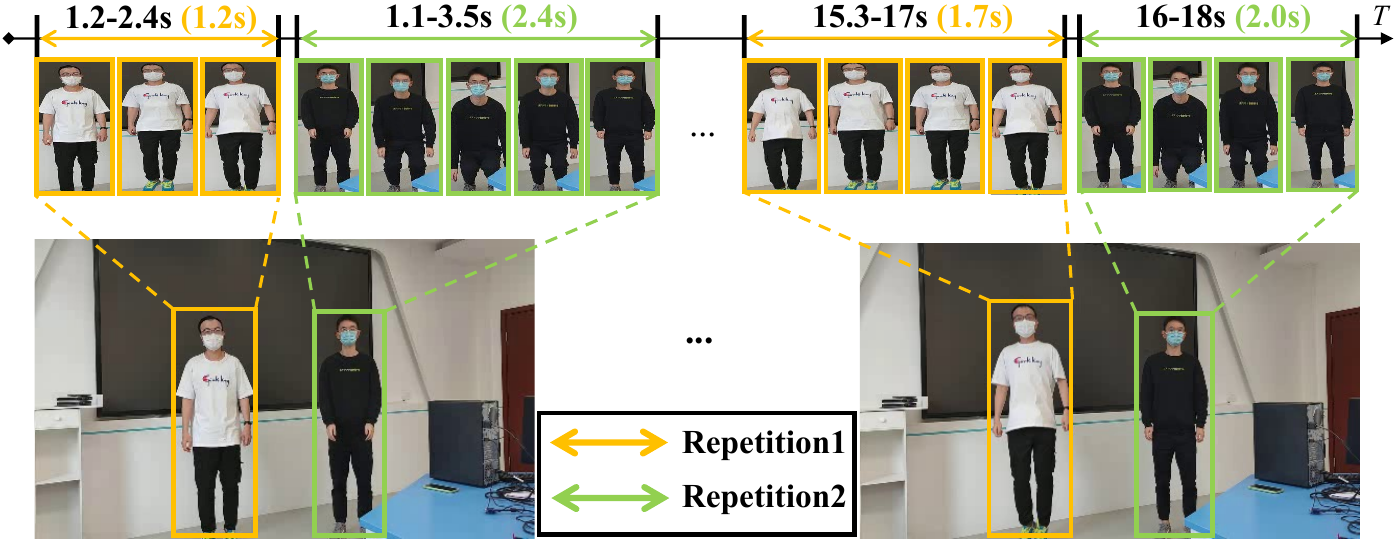}
\caption{\textbf{The illustration on MRAC tasks in untrimmed videos.} In this example, two people are performing different repetitive actions asynchronously, with variable cycle lengths.}
\label{fig:intro}
\end{figure}

Multi-instance Repetitive Action Counting (MRAC) has emerged as a computer vision task that aims to accurately count the number of repetitive actions performed by multiple instances in untrimmed videos. As shown in Figure \ref{fig:intro}, periodic activities in untrimmed videos often feature with variety and asynchronism. These repetitions may not keep a consistent velocity and could start mid-sequence even containing occasional pauses.
An efficient MRAC system holds significant importance in conducting comprehensive analysis of human-centric activities, including applications such as AI-powered sports action analysis \cite{hanyao2021edge,xie2023actor} and healthcare monitoring \cite{alnaggar2023video}.
Progress has been made for the Single Repetition Action Counting (SRAC) task, which only involves repetitive actions of one instance \cite{dwibedi2020counting,hu2022transrac,yao2023poserac}.
Particularly, PoseRAC \cite{yao2023poserac} improves single-person repetition counting through pose estimation, but it cannot recognize unseen actions during training such that fails to generalize to open-set scenarios.
However, handling real-world settings where multiple instances engage in simultaneous repetitive actions remains a challenge.

Integrating a tracker with the SRAC approach is inefficient for MRAC tasks. Focusing on instances themselves loses global information representation. Running the SRAC models separately for each instance is time-consuming with multiple instances. 
For this limitation, we propose MultiCounter, an end-to-end deep learning framework specifically designed to detect, track, and count repetitive actions of multiple human instances in untrimmed videos. To the best of our knowledge, we formally propose the MRAC task for the first time. It involves three significant challenges: 
1) MRAC requires effective spatiotemporal modeling to distinguish the unique repetitive action patterns of each instance across consecutive frames.
2) Accurately determining the temporal boundaries of numerous action-agnostic repetitions at the instance level in untrimmed videos, and then counting the repetitions of each instance is inherently difficult.
3) The lack of MRAC-specific datasets and evaluation metrics prevents the task practical.

To tackle this, we provide a unified solution to address the potential challenges inherent in MRAC tasks, including the following aspects: \textbf{1) Efficient Spatiotemporal Correlation.} We propose the Mixed Temporal-Spatial Interaction (MSTI) module to effectively model spatiotemporal correlations. The MSTI module uses instance-specific queries and multi-scale features for instance-level spatiotemporal modeling.
This enables MultiCounter to handle complex multi-instance repetitive actions.
\textbf{2) Action-agnostic Repetition Counting.} We further propose two task-specific heads: the Instance Head and Period Head, aiming to locate all potential human instances and predict their periodic properties. 
By robustly detecting and localizing temporal boundaries of multiple action-agnostic repetitions, our approach generalizes to action categories not seen during training.
\textbf{3) MRAC Benchmark.} To evaluate the MRAC task, we synthesize a dataset called MultiRep based on existing SRAC benchmarks to train MultiCounter. Additionally, we propose a new evaluation metric called Period-AP to reflect periodic localization ability within each repetitive action across all frames.

In contrast to merely combining a tracker with SRAC methods, MultiCounter is a fully end-to-end framework that simultaneously detects multiple human instances, tracks them over time, and counts the number of repetitive actions. Different from Transformer-based models like DETR \cite{carion2020end}, MultiCounter captures spatial-temporal features related to the “action of interest (AoI)” across consecutive frames. This AoI-wise design eliminates redundancies and enforces MultiCounter to focus more on informative regions. On the contrary, DETR only models spatial context using frame-wise dense interaction, making it incapable for MRAC tasks. DETR is also rooted in the closed-set assumption that the test set only contains the pre-defined object categories. It cannot handle the inevitable unknown actions in open-set scenarios. We conduct in-depth experiments by training MultiCounter on the MultiRep dataset generated from a labeled real-world RAC benchmark, validating the key challenge of counting multiple repetitive actions in untrimmed videos and further showcasing the superiority of our proposed method. The main contributions are as follows:
\begin{itemize} 
    \item We formally define and explore multi-instance repetitive action counting in untrimmed videos, and propose a fully end-to-end deep learning framework called MultiCounter, which establishes a new benchmark in the field of MRAC. 
    \item To model complex spatiotemporal correlations across consecutive frames, we design the Mixed Spatial-Temporal Interaction (MSTI) module and propose two task-specific heads to accurately output temporal boundaries of multiple action-agnostic repetitions. 
    \item We synthesize an MRAC dataset termed MultiRep to train MultiCounter and propose a new metric called Period-AP for performance evaluation.
    \item Experiments on the MultiRep dataset show that MultiCounter improves Period-mAP by 41.0\%, reduces AvgMAE by 58.6\%, and increases AvgOBO 1.48 times when compared to ByteTrack+RepNet. MultiCounter also achieves real-time MRAC regardless of the number of human instances present.  
    
\end{itemize}


\section{Related Work}
\subsection{Action Spatial-temporal Modeling}
Action spatial-temporal modeling refers to the process of understanding and representing both the spatial and temporal aspects of human actions in videos, mainly including video action recognition, and temporal action localization. For the former, convolutional neural networks (CNNs) based methods \cite{tran2015learning,feichtenhofer2016convolutional,carreira2017quo,qiu2017learning} dominate for a long time. Considering the local correlation nature of CNNs, which cannot capture long-term dependencies across temporal domains, recurrent networks such as RPAN \cite{du2017rpan}, MRRN \cite{zheng2017multi}, and CRN \cite{sun2018coupled} have been proposed successively. To enlarge the receptive field effectively and obtain global feature representation, recent works focus on Transformer-based pipelines like ViT \cite{bertasius2021space} and its several variants \cite{patrick2021keeping,liu2022video}, which makes accurate action recognition possible. Temporal action localization aims to classify and temporally localize each action instance in untrimmed videos. Most of the existing methods have fallen into the one-stage \cite{long2019gaussian,liu2022end,zhao2022tuber} or two-stage paradigms \cite{escorcia2016daps,buch2017sst,chao2018rethinking}. One-stage approaches can simultaneously recognize and localize actions, typically training in an end-to-end manner without human detectors or proposals. The two-stage methods initially generate multiple spatiotemporal proposals, followed by individual classification and boundary refinement for each proposal. However, these works are suitable for the spatiotemporal modeling of known action categories. MultiCounter aims to achieve the accurate perception of numerous action-agnostic instances across consecutive frames.
\begin{figure*}[htbp]
		\hspace*{0cm}
		\centering
		\includegraphics[scale=0.8]{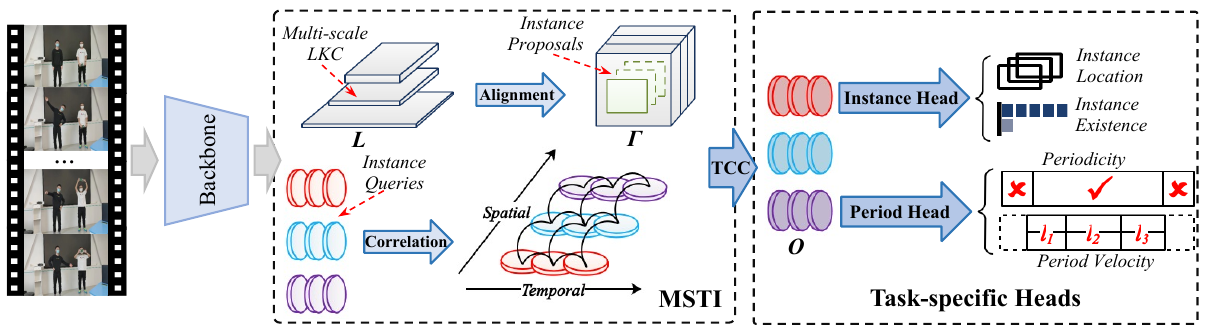}
		\caption{\label{Fig2}\textbf{Overview of the MultiCounter framework.} It simultaneously detects multiple human instances, tracks them over time, and counts the number of repetitive actions. This paradigm helps facilitate joint optimization of sub-tasks by updating $M$ times of instance queries and proposal boxes iteratively at every backpropagation. \textbf{LKC}: Large-Kernel Convolution, \textbf{TCC}: Temporal Conditional Convolution, \textbf{\emph{L}}: Latent features, \textbf{$\Gamma$}: RoI features, \textbf{\emph{O}}: Target features.}
		\label{fig:model}
\end{figure*}
\subsection{Repetition Action Counting}
Earlier repetition action counting methods rely on signal processing techniques such as Fourier analysis \cite{pogalin2008visual}, wavelet transform, and peak detection to transform periodic motion patterns into one-dimensional signals and count the number of repetitions \cite{cutler2000robust}. Recently, the deep learning-based model has been presented by \cite{levy2015live}, which enables estimating the number of repetitive actions in a video by training efficient CNNs on the synthetic dataset. \cite{runia2018real} further proposes to analyze motion fields based on gradient, curl, and divergence, which can well solve the periodic change problem. \cite{zhang2020context} trains a context-aware and scale-insensitive framework leveraging 3D CNNs on a novel single-action repetition counting benchmark derived from UCF101 \cite{soomro2012ucf101}. Meanwhile, \cite{dwibedi2020counting} proposes a novel deep-learning method for temporal repetition counting. By training it with synthetic repetition videos, the proposed model enables good generalization in real-world scenarios. Unlike works that only use a visual modality, \cite{zhang2021repetitive} incorporate the sound into the repetition action counting tasks. Such a scheme of multiple modalities collaboration can well compensate each other, thereby improving the counting result of the model. More recently, a full-resolution repetition counting model has been proposed to comprehensively understand repetitive actions from a complete temporal perspective \cite{li2023full}. \cite{hu2022transrac} introduces a large-scale RepCount dataset featuring fine-grained annotations of actions, and proposes a multi-scale temporal transformer model to achieve precise repetition counting. To balance effectiveness and efficiency, \cite{yao2023poserac} proposes a lightweight pose-level repetition counting method where only salient poses and actions are used for training. However, they are designed specifically for SRAC tasks, thus cannot directly applied to MRAC scenarios.


\section{Task Definition}
Unlike the case of single repetition, MRAC involves the task of identifying and localizing multiple repetitive actions within given untrimmed videos and then counting the number of each repetition. The task requires algorithms to analyze the temporal dynamics of multiple instances and accurately pinpoint the periodic temporal boundaries at the instance level. This highlights the challenges of MRAC tasks in long videos.

Given an untrimmed video containing $T$ frames, the MRAC algorithm predicts $K$ human instances of performing repetitive actions. For the $k$-th person in the video, this algorithm outputs a triplet $Y^{K}_{k=1} =\{ L_k, P_k, C_k\}$. In specific, $L$ indicates the instance-level repetition action location \emph{i.e.,} bounding boxes. If the $k$-th human instance of the $t$-th frame is not visible (\emph{e.g.,} due to the serious occlusion or temporal absence), then $L^{k}_{t}$ is set to $\emptyset$. $P = [s, e]$ is the temporal boundaries of instance-level repetition action, namely repetition proposals in the following, where $s$ and $e$ denote the start and end points of a repetitive action. The value of $C$ denotes the final count. For the ground truth of an MRAC task, we add a tilde to the corresponding notation, such that we have $\tilde{Y}^{Z}_{z=1} =\{ \tilde{L}_z, \tilde{P}_z, \tilde{C}_z\}$ to denote the ground truth triplet, where $Z$ is the number of human instances in the video. The good performance is achieved when the prediction $Y^{K}_{k=1}$ is closer to the ground truth $\tilde{Y}^{Z}_{z=1}$.


\section{Model}
In this section, we propose the design of MultiCounter architecture. It takes a video clip as input and outputs the predictions of each human instance in the whole clip. MultiCounter runs in three steps. Firstly, MultiCounter feeds the video clip to a visual backbone to obtain the multi-scale feature embeddings. Then, the mixed spatial-temporal interaction (MSTI, $\S$ \ref{Mixed Spatial-Temporal Interaction}) module models the spatial-temporal context interaction of feature embeddings to get target features. Finally, the task-specific heads ($\S$ \ref{Task-specific Heads}) perceive the temporal boundaries from the target features to output the instance-level predictions. 

For efficient training of MultiCounter, the predictions consist of human existence classification $D$, action location $L$, period velocity $\psi$, and periodicity $\varphi$. Particularly, we set $\psi$ = $e$-$s$ as the rate at which repetitive actions occur, and $\varphi$ denotes whether the frame is within a repetition proposal of the clip. The framework of MultiCounter is shown in Figure \ref{fig:model}. Considering a video clip $V=\left[ v_{1},v_{2},...,v_{t}\right]  \in \mathbb{R}^{T\times 3\times H\times W}$ as a continuous sequence of frames, where $T$ represents the total number of frames, and $3\times H\times W$ denotes the spatial dimensions of each RGB image. Firstly, we feed the video clip $V$ to a visual backbone $B$ to obtain multi-layer feature pyramid embeddings $F=\left[ f_{1},f_{2},...,f_{n}\right]$ in which $f_{i}\in \mathbb{R}^{T\times C^{\prime }\times H^{\prime }\times W^{\prime }}$, $C^{\prime }$, $H^{\prime }$, $W^{\prime }$ are the number of channels, height of the feature, and width of the feature, respectively. Next, the MSTI module is applied to $F$ to obtain the latent features. We further employ an instance query-based sub-architecture \cite{sun2021sparse,zhao2022tuber} to perform sparse interaction between latent features and query features iteratively. After each interaction, instance queries will be updated and the results (\emph{i.e.,} instance-level $D$, $L$, $\psi$, $\varphi$) are finally output by task-specific heads of the last iteration.

\subsection{Mixed Spatial-Temporal Interaction (MSTI)}
\label{Mixed Spatial-Temporal Interaction}
To effectively model spatiotemporal correlation of multi-instance actions across consecutive frames, we design the MSTI module. Specifically, MSTI first employs a large-kernel convolution submodule behind each feature pyramid embedding \cite{ding2022scaling,guo2023visual} to capture local-global dependencies, which consists of a local depth-wise convolution across multi-instance actions, a long-range dilation convolution along temporal frames, and a $1 \times1$ vanilla convolution. Finally, we achieve the multi-scale context interaction across all layers to get latent features in a cascaded manner. We also apply RoI alignment \cite{he2017mask} to the latent features to obtain instance-specific RoI features $\Gamma \in \mathbb{R}^{T\times C^{\prime \prime}\times H^{\prime \prime}\times W^{\prime \prime}}$. It helps MultiCounter understand contextual information that contains multiple different repetitive actions.

After obtaining instance-specific RoI features $\Gamma$, MSTI leverages a set of instance queries $\left\lfloor Q\right\rfloor  \in \mathbb{R}^{N\times T\times C}$ (\emph{i.e.,} $N$ spatio-temporal embeddings) and instance proposals $\left\lfloor P\right\rfloor  \in \mathbb{R}^{N\times T\times 4}$ (\emph{i.e.,} $N$ spatio-temporal bounding boxes) pairs to generate sparse candidates, with each pair of candidates representing one human instance across the whole video. To effectively associate different queries with the corresponding instances and ultimately model the representation of their motion patterns, we first use a temporal block to achieve communication specific to instances across all frames:
\begin{equation}
\small
\begin{aligned}
\left\lfloor Q\right\rfloor^{T}_{t=1}  =\left\lfloor Q\right\rfloor^{T}_{t=1}  +Norm\left( MHSA\left( \left\lfloor Q\right\rfloor^{T}_{t=1}  \right)  \right)  ,
\  i\in \left[ 1,\  N\right]  
\end{aligned}
\end{equation} 
where $MHSA$ and $Norm$ denote multi-head self-attention \cite{vaswani2017attention} and layer normalization \cite{ba2016layer}, respectively. After obtaining temporal association, a spatial block is then employed to facilitate spatial interaction among all instance queries within each frame and eventually get the query embeddings:
\begin{equation}
\small
\begin{aligned}
\left\lfloor Q\right\rfloor^{N}_{i=1}  =\left\lfloor Q\right\rfloor^{N}_{i=1}  +Norm\left( MHSA\left( \left\lfloor Q\right\rfloor^{N}_{i=1}  \right)  \right)  ,
\  i\in \left[ 1,\  T\right]
\end{aligned}
\end{equation}


Based on the RoI features $\Gamma$ and query embeddings after spatiotemporal correlation using instance queries, we design the Temporal Conditional Convolution (TCC) module to get target features. Specifically, we generate a dynamic convolution filter \cite{yang2019condconv,chen2020dynamic} conditioned on the query embeddings and then use it to perform 3D convolution on the $\Gamma$ to get filtered features. The filtered features are projected linearly by a fully connected layer to get the final target features $O \in \mathbb{R}^{N\times T\times C}$. As the number of iterations increases, MSTI achieves highly efficient instance-level spatiotemporal context modeling, resulting in discriminative representations of action positions and the periodic properties of each instance.

\subsection{Task-specific Heads}
\label{Task-specific Heads}
To achieve accurate perception of the human instance location and get the periodic properties from target features, we design two task-specific heads: instance head and period head. The former is responsible for precisely recognizing and localizing all possible human instances, while the latter is used to obtain the per-frame period velocity and periodicity of all instances, which are bound to repetition proposals and counts across the video clip.

\textbf{Instance Head.} Given the target features, we use two multi-layer perceptron (MLP) layers to detect and locate the instances:
\begin{equation}
\small
D=Sigmoid\left( MLP_{d}\left( O\right)  \right), L=MLP_{l}\left( O\right)  
\end{equation}
where $D\in \mathbb{R}^{N\times T}$ denotes instance existence or not and $L\in \mathbb{R}^{N\times T\times 4}$ is the action bounding boxes across the whole clip. We also use $L$ to renew the instance proposals $\left\lfloor P\right\rfloor$ at the next iteration.

\textbf{Period Head.} 
To enable the well-trained model to infer the unseen repetitive actions during training, we design the cross-temporal dual-attention mechanism (CDM) into the prediction of the period head. After obtaining the updated target features $O$ for each instance across the whole frames, we begin by constructing an inter-frame self-similarity matrix through the calculation of all pairwise feature similarities $S$ and intra-frame temporal correlations by computing attention scores $A$:
\begin{equation}
\small
S=ReLU\left( Softmax\left( \frac{cs\left( O,O^{T}\right)  }{\sqrt{dim} } \right)  \right) 
\end{equation}
\begin{equation}
\small
A=Attention\left( \left\lfloor Q_{O}\right\rfloor  ,\left\lfloor K_{O}\right\rfloor  \right)
\end{equation}
where $cs\left( O, O^{T}\right)$ denotes the cosine distance between pairwise frames, $dim$ is the number of channels, $ReLU$ is the $ReLU$ activation,  $Attention(.)$ denotes the dot product attention, $\left\lfloor Q_{O}\right\rfloor$ and $\left\lfloor K_{O}\right\rfloor$ are the query and key matrix, respectively. Next, we apply a Transformer block with $3 \times3$ convolution into the embeddings obtained by concatenating $S$ and $A$ to learn this latent rule that implies salient periodic motion patterns. This enables robust action-agnostic generalization through continuous iterative training. After performing CDM module to get $O^{\prime }\in \mathbb{R}^{N\times T\times C}$, we use the period head to achieve period velocity estimation and periodicity detection:
\begin{equation}
\small
\psi=Softmax\left( MLP_{\psi}\left( O^{\prime }\right)  \right), \varphi =Sigmoid\left( MLP_{\varphi }\left( O^{\prime }\right)  \right) 
\end{equation}
where $\psi\in \mathbb{R}^{N\times T\times \frac{T}{2} }$ denotes per-frame period velocity classification score and $\varphi \in \mathbb{R}^{N\times T }$ indicates per-frame binary periodicity classification. We treat period velocity prediction as a classification task, that is, use discrete ground truth for supervised training. Taking into account the nature of the repetition, the maximum period velocity is half the number of given input frames.

\subsection{Loss}
For each video clip, we consider the output of both the instance and period heads to calculate the total loss of MultiCounter. This involves initially establishing a one-to-one assignment between instance-level predictions and ground truths through bipartite matching. Based on this assignment, we calculate the training loss. Given ground truths $\tilde{Y}_{Z} =\left\{ \left( \tilde{D},\tilde{L},\tilde{\psi},\tilde{\varphi} \right)  \right\}  $, where $\tilde{D} \in \mathbb{R}^{Z\times T}$, $\tilde{L} \in \mathbb{R}^{Z\times T\times 4}$, $\tilde{\psi} \in \mathbb{R}^{Z\times T\times \frac{T}{2}}$, $\tilde{\varphi} \in \mathbb{R}^{Z\times T}$ represent the existence of human instances, action bounding boxes, action period velocity classification score and binary periodicity classification, respectively. Inspired by the previous works \cite{carion2020end,sun2021sparse}, we employ the Hungarian algorithm to conduct bipartite matching between predictions and ground truths:
\begin{equation}
\small
\begin{aligned}
L_{hung}\left( \tilde{Y}_{Z} ,Y_{K}\right)  =L_{cls}\left( \tilde{D} ,D\right)  + 1_{\tilde{D} \neq 0}\left( L_{bbox}\left( \tilde{L} ,L\right)  \right)  
\end{aligned}
\end{equation}
here $L_{cls}\left( \tilde{D}, D\right)$ denotes the sigmoid focal loss for instances existence detection while $L_{bbox}\left( \tilde{L}, L\right)$ is a combination of $L1$ loss and $GIoU$ loss for locating the position of different instances. After obtaining the optimal assignment $\sigma $ between the prediction of the instance head and corresponding ground truths, we perform network optimization. The matched loss function is computed as:
\begin{equation}
\small
\begin{aligned}
L_{total}=L_{inst}\left( \tilde{Y} ,\tilde{Y}^{\sigma } \right)  +\lambda L_{period}\left( \tilde{\psi} ,\tilde{\psi}^{\sigma } \right)  + \\ 
\mu L_{periodicity}\left( \tilde{\varphi} ,\tilde{\varphi}^{\sigma } \right)  
\end{aligned}
\end{equation}
where $L_{inst}\left( \tilde{Y},\tilde{Y}^{\sigma } \right)$ indicates the loss that has already been matched human instances between predictions and ground truths, which shares the same loss functions as $L_{hung}(.)$. Similarly, $L_{period}\left( \tilde{\psi}, \tilde{\psi}^{\sigma} \right)$ and $L_{periodicity}\left( \tilde{\varphi},\tilde{\varphi}^{\sigma} \right)$ are the loss of period velocity as well as periodicity and are optimized using multi-class cross-entropy and binary cross-entropy, respectively. In our experiment, $\lambda$ and $\mu$ are set to 1 and 5, respectively.


\section{Experiment}
Counting multiple repetitive actions in untrimmed videos has rarely been explored before. Thus, we introduce a synthetic dataset and corresponding evaluation metrics to evaluate MultiCounter.
\begin{table*}
\caption{\textbf{Main results on the MultiRep dataset.} ‘--’: Failure to provide frame-level results. ‘$T$(=49.4ms)’: Time consumption for human instance tracking. ‘$\dag$’: Time consumption for pose pre-extraction and inference. ‘instances$^{\ast }$’: Human instance amount in the given frame.}
\renewcommand\arraystretch{1.2}
\centering
\setlength{\tabcolsep}{1.5mm}{
    \begin{tabular}{lccccccc}
        \bottomrule
        Method  & Period-mAP & Period-AP$_{50}$ & Period-AP$_{75}$ & AvgMAE $\downarrow $ & AvgOBO $\uparrow $  & Params (M) & Time/frame (ms) \\
        \hline
        ByteTrack \cite{zhang2022bytetrack}+RepNet \cite{dwibedi2020counting}      & 7.52    & 26.16      & 3.11     & 0.577 &0.223 & 112.7&$T$(=49.4)+12.7$\times$instances$^{\ast }$ \\
        ByteTrack \cite{zhang2022bytetrack}+TransRAC \cite{hu2022transrac}    & 2.26     & 8.78     & 0.87     & 0.542 &0.267 & 110.5 &$T$+24.3$\times$instances$^{\ast }$  \\
        ByteTrack \cite{zhang2022bytetrack}+PoseRAC \cite{yao2023poserac}      & --    & --      & --     & 0.361 & 0.531  & \textbf{95.1} & $T$+(39.1$^\dag $)$\times$instances$^{\ast }$ \\
        \hline
        \textbf{MultiCounter (Ours)} & \textbf{10.60}  & \textbf{34.25}     & \textbf{5.40}   & \textbf{0.239} & \textbf{0.554} & 113.6  & \textbf{15.9}(data processing)+\textbf{23.7} \\
        \toprule
    \end{tabular}}
    \label{tab:MultiRep}
\end{table*}

\subsection{Dataset}
We synthesize data to train our MultiCounter based on existing SRAC datasets that capture single-action repetitions. We generate $1,157$ synthetic videos with $52,590$ periodic events to get the synthetic dataset called MultiRep and split these into three sets in a ratio of $7:2:1$ for training, validation, and testing.

\begin{figure}[htbp]
		\hspace*{0cm}
 		\centering
 		\includegraphics[scale=0.33]{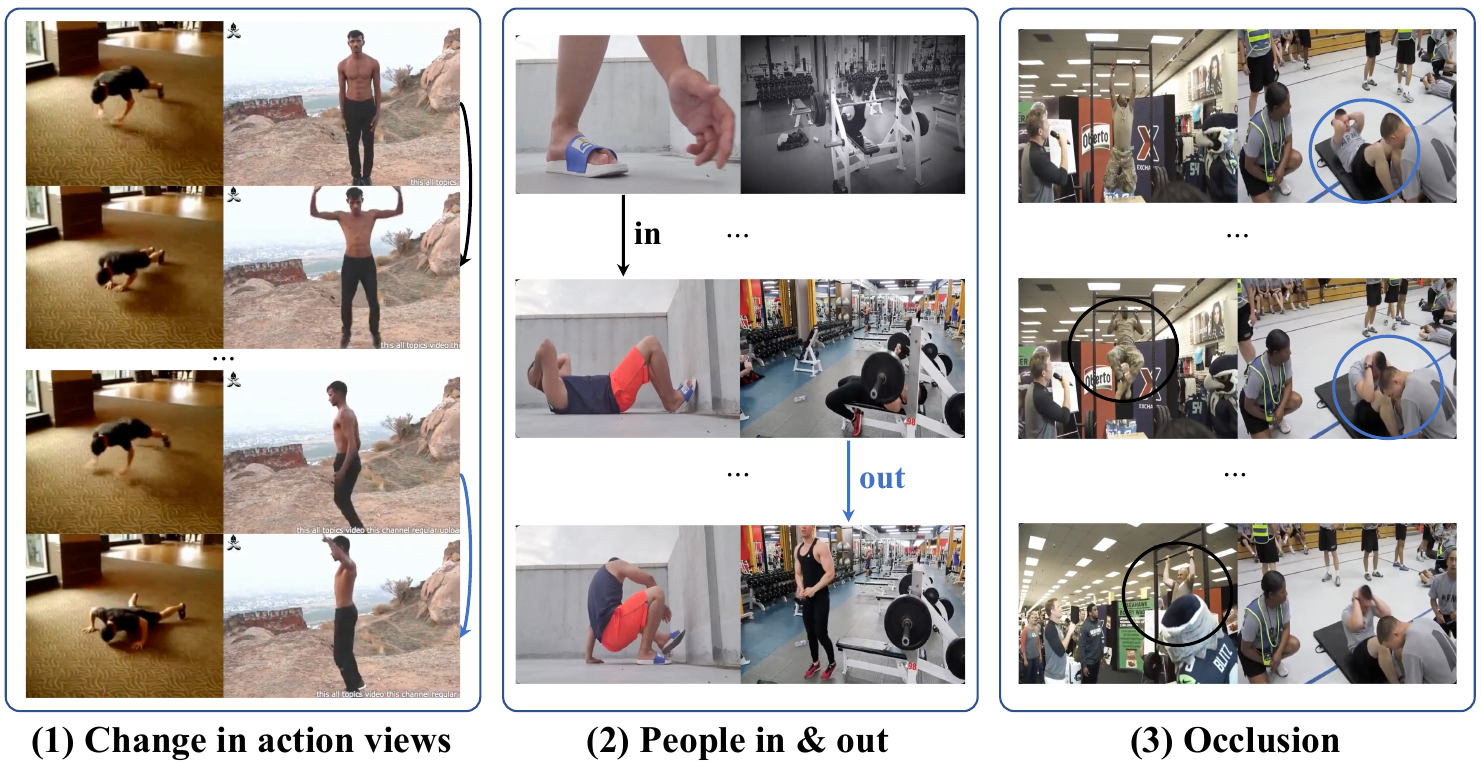}
 		\caption{\label{Fig2} Data samples in our MultiRep synthetic dataset.}
		\label{fig:data}
\end{figure}

\textbf{Video Generation.} 
We group all samples from the RepCount dataset \cite{hu2022transrac} according to frames per second (FPS). We randomly select two or three untrimmed videos from the same FPS candidates until all samples have been selected at least once. Except for the number of candidate instances, there are no differences in synthesizing videos with two and three instances. In our synthetic dataset, the number of data samples with two and three instances are $643$ and $514$, respectively. Our way of synthesizing videos can easily extend to more instances. We further check and eliminate duplicate cases, such as $A+B+C$ and $B+C+A$. Finally, we use the FFmpeg video processing tool to horizontally concatenate these untrimmed videos. Note that this does not introduce the position changes of instances or occlusion. However, as shown in Figure \ref{fig:data}, there are various clutters in our dataset like changes in action views over time, people in \& out, and occlusion, which highlights significant challenges for MRAC tasks. We also use RandomFlip and Scale Transformation for data augmentation.

\textbf{Data Annotation.} 
We label bounding boxes of each human instance performing repetitive actions across the video. Technically, we use YOLOv8 with the ByteTrack tracker \cite{zhang2022bytetrack} for semi-supervised annotation. Two human annotators are responsible for rigorous inspection and correction of the tracking results. For periodic annotation, we use the fine-grained ground truths of RepCount dataset, which marks the start and end frame point of each repetitive action. For the period velocity ground truth of each repetition, we get it by subtracting the beginning timestamp from the end timestamp. Accordingly, video frames within the repeating segments are assigned a value of 1 to indicate their periodicity ground truth, otherwise 0.

\subsection{Evaluation Metrics}
Since MultiCounter outputs per-frame action period velocity and periodicity at the instance level rather than the number of repetitions, we introduce a new metric termed Period-AP to evaluate the periodic localization ability of the model across all frames. Meanwhile, existing SRAC metrics \cite{zhang2021repetitive,hu2022transrac,yao2023poserac} are modified to fit MRAC tasks.

\textbf{Period-AP.} 
Taking inspiration from temporal action localization and detection \cite{kalogeiton2017action}, we tailor its original AP metric into our task to reflect MultiCounter's periodic localization ability within each repetitive action. Based on the predictions of instance-level action period velocity and periodicity, we get the repetition proposal of each instance by setting the start and end anchors. Next, we calculate the 3D segment Intersection over Union (sIoU) across all frames based on matched pairs with true positive (TP) predictions and ground truths. We then report Period-AP under the temporal sIoU of 50\%, 75\%, and 50\%-95\% with step 5\%.

\textbf{Avg-OBO and Avg-MAE.}
Building upon the SRAC metrics MAE (Mean Absolute Error) and OBO (Off-By-One count accuracy), we extend them to MRAC tasks and introduce two new metrics called Avg-MAE and Avg-OBO. We get the predictions with period velocity as well as periodicity and count repetitions only within the repetitive segments, that is, $C=\sum^{T}_{t=1} c_{t}$, where $c_{t}=\frac{1}{\psi_{t}}\left( \varphi_{t} \neq 0\right)$ and $T$ denote per-frame valid period velocity (\emph{i.e.,} the period velocity within repeating segments) and the number of frames, respectively. Accordingly, AvgMAE and AvgOBO can be defined as:
\begin{equation}
\small
AvgMAE=\frac{1}{N^{\diamond}} \sum^{N^{\diamond}}_{i=1} \left( \frac{1}{M^{\diamond}} \sum^{M^{\diamond}}_{j=1} \frac{\tilde{C}_{i,j} -C_{i,j}}{\tilde{C}_{i,j} } \right)  
\end{equation}
\begin{equation}
\small
AvgOBO=\frac{1}{N^{\diamond}} \sum^{N^{\diamond}}_{i=1} \left[ \left( \frac{1}{M^{\diamond}} \sum^{M^{\diamond}}_{j=1} \left| \tilde{C}_{i,j} -C_{i,j}\right|  \leq 1\right)  \right]  
\end{equation}
where $N^{\diamond}$ and $M^{\diamond}$ are the number of videos and human instance amounts with repetitive actions within a video, respectively. $\tilde{C}_{i,j}$ and $C_{i,j}$ denote ground truth and prediction count.

\subsection{Implementation Details}
We develop MultiCounter based on PyTorch. We use ResNet-50-FPN \cite{lin2017feature} pre-trained on ImageNet-1K as the backbone to extract base features. We sample the consecutive frames with different rates such that each clip covers at least two repetitions. The clip length and image size are set to 64 and $224\times224$, respectively. We set the number of queries and iteration times to 20 and 4, respectively. We use NVIDIA Tesla V100-PCIE with 32GB GPUs for model training and Adam with a batch size of 4 for speedup. The training process spans $10,000$ iterations, with the learning rate being reduced by a factor of 0.1 at iterations $4,000$ and $7,000$, starting from an initial learning rate of $6e-6$. During inference, we use 64 consecutive windows with an overlap of 32 and feed it as input to MultiCounter, which outputs per-frame period velocity and periodicity at the instance level. Moreover, predictions between adjacent clips are linked by the IoU scores of human instances bounding boxes. These default settings are used for all ablation studies unless otherwise specified.

\subsection{Main Results}
\textbf{Performance on MultiRep Dataset.} To comprehensively evaluate the superiority of MuiltCounter, we customize several representative SRAC methods \cite{dwibedi2020counting,hu2022transrac,yao2023poserac} using a unified instance tracking solution and provide the performance comparison on MultiRep Dataset. To be specific, we use Bytetrack \cite{zhang2022bytetrack} to perform per-human instance detection and tracking, ultimately obtaining a sequence of action bounding boxes by IoU similarity matching between adjacent frames. Next, we employ three mainstream RAC algorithms within each action tracklet (\emph{i.e.,} instance-level action bounding boxes across all frames) and report their Period-AP, AvgMAE and AvgOBO metrics. We also show the inference speed comparison on a single NVIDIA Tesla V100 GPU. The results can be seen in Table \ref{tab:MultiRep}.

Firstly, MultiCounter has a higher Period-AP. This is because our approach can effectively model local-global context to locate repetition proposals of multiple human instances, whereas the tracking-by-detection framework lacks global information representation, particularly in the presence of interference such as occlusion and departure. Secondly, MultiCounter achieves significant performance gains in AvgMAE and AvgOBO. The results are attributed to the proposed joint optimization framework in MultiCounter, which benefits the sub-tasks of detection, tracking, and counting uniformly. In other words, the proposed model effectively captures spatio-temporal correlations of periodic patterns at the instance level, leading to robust MRAC. Finally, as evidenced in Table \ref{tab:MultiRep}, MultiCounter exhibits a higher inference speed even with a marginal rise in model parameters against alternative methods. This advantage stems from MultiCounter's one-stage inference process, whereas the runtime of counterparts increases linearly with the number of instances.
\begin{table}[]
\caption{\textbf{Results on the RepCount and UCFRep datasets.} MultiCounter is trained on MultiRep and tested on RepCount and UCFRep.}
\renewcommand\arraystretch{1.2}
\centering
\setlength{\tabcolsep}{2mm}{
\begin{tabular}{l|cc|cc}
\bottomrule
\multicolumn{1}{l|}{\multirow{2}{*}{Method}} & \multicolumn{2}{c|}{RepCount}      & \multicolumn{2}{c}{UCFRep}        \\ \cline{2-5} 
\multicolumn{1}{c|}{}                        & \multicolumn{1}{l|}{MAE $\downarrow $}   & OBO $\uparrow $   & \multicolumn{1}{l|}{MAE $\downarrow $}   & OBO $\uparrow $   \\ \hline
RepNet \cite{dwibedi2020counting}                                      & \multicolumn{1}{c|}{0.995} & 0.013 & \multicolumn{1}{c|}{0.999} & 0.009 \\ 
Zhang et al. \cite{zhang2020context}                                    & \multicolumn{1}{c|}{0.879} & 0.155 & \multicolumn{1}{c|}{0.629} & 0.297 \\ 
TANet \cite{liu2021tam}                                     & \multicolumn{1}{c|}{0.662} & 0.099 & \multicolumn{1}{c|}{0.892} & 0.129 \\ 
Video-SwinT \cite{liu2022video}                                   & \multicolumn{1}{c|}{0.576} & 0.132 & \multicolumn{1}{c|}{1.122} & 0.033 \\ 
Huang et al. \cite{huang2020improving}                                & \multicolumn{1}{c|}{0.527} & 0.159 & \multicolumn{1}{c|}{1.035} & 0.015 \\ 
TransRAC \cite{hu2022transrac}                                    & \multicolumn{1}{c|}{0.443} & 0.291 & \multicolumn{1}{c|}{0.640} & 0.324 \\ 
PoseRAC \cite{yao2023poserac}                                       & \multicolumn{1}{c|}{0.236} & \textbf{0.560} & \multicolumn{1}{c|}{0.638} & 0.229 \\ \hline
\textbf{MultiCounter (Ours)}                           & \multicolumn{1}{c|}{\textbf{0.232}}      &     0.527  & \multicolumn{1}{c|}{\textbf{0.432}}      & \textbf{0.343}      \\
\toprule
\end{tabular}}
\label{tab:RepUCF}
\end{table}
    
\textbf{Cross-dataset Generalization.} To verify MultiCounter’s generalization on SRAC task, we conduct experiments on both RepCount \cite{hu2022transrac} and UCFRep \cite{zhang2020context} benchmarks and report the mainstream evaluation metrics, including MAE and OBO. As shown in Table \ref{tab:RepUCF}, MultiCounter trained on MultiRep and tested on both RepCount and UCFRep outperforms most counterparts, highlighting its cross-dataset generalization ability. Although PoseRAC \cite{yao2023poserac} provides slight OBO gains on RepCount, which benefits from its strong feature representation of known actions. MultiCounter exceeds it significantly on UCFRep which includes several repetitive action categories not seen during training (\emph{e.g.,}  Rowing and SoccerJuggling). This indicates that our model also enables generalization to action-agnostic SRAC tasks.

\subsection{Ablation Studies}
\textbf{Effect of Submodules.} We perform several ablation studies to demonstrate the effect of the proposed modules on enhancing MultiCounter's performance. As the results shown in Table \ref{tab:abs}, the absence of any submodules adversely impacts model performance. For the MSTI module, adding multi-scale context information interaction improves both AvgMAE and AvgOBO by 0.05 when comparing with the counterpart used base features only, indicating the long-term dependencies representation ability of it. We can also observe that AvgMAE increases from 0.24 to 0.41, AvgOBO drops from 0.55 to 0.35, and Period-mAP decreases by 8.24 without the use of temporal blocks, highlighting the critical importance of temporal correlation for periodic detection and localization in untrimmed videos. Similarly, applying spatial correlation also boosts model’s performance, since the spatial interaction facilitates information exchange among queries to be beneficial to modeling specific instance features. Accordingly, the Period-mAP, AvgMAE, and AvgOBO all have negative effects if directly feeding instance-specific RoI features into fully connected layers and renewing queries. This is because the TCC module leverages convolutional weights to selectively gather important instance features while filtering out irrelevant ones or background, contributing to discriminative representations of task-specific heads. The CDM module substantially improves model performance, especially Period-mAP. It also helps in generalizing to unknown repetitive actions during training, which is in line with results on the UCFRep dataset.
\begin{table}[]
\caption{\textbf{Ablation studies of different modules.}}
\renewcommand\arraystretch{1.2}
\centering
\setlength{\tabcolsep}{2mm}{
\begin{tabular}{lccc}
\bottomrule
Method             & Period-mAP & AvgMAE $\downarrow $ & AvgOBO $\uparrow $ \\
\hline
w/o context interaction           &   9.03      & 0.29     & 0.50     \\
w/o temporal block & 2.36        & 0.41     & 0.35     \\
w/o spatial block  & 10.21        &  0.31      & 0.45    \\
w/o TCC            & 8.64        & 0.29     & 0.49     \\
w/o CDM            & 7.96        & 0.32     & 0.46     \\
\hline
\textbf{full model}         & \textbf{10.60}     & \textbf{0.24}   & \textbf{0.55}  \\ 
\toprule
\end{tabular}}
\label{tab:abs}
\end{table}

\textbf{Convergence Speed.} 
In the design of MultiCounter, we collect the “action of interest (AoI)” features using a set of instance queries and proposals. We further evaluate the effectiveness of the context interaction module by replacing it with the existing Transformer encoder in VisTR \cite{wang2021end}. As shown in Table \ref{tab:cs}, compared to the alternative that uses frame-wise dense interaction in VisTR, MultiCounter eliminates backgrounds and redundancies by such AoI-wise design, achieving 12.5$\times$ speed up of convergence. Benefiting from the AoI-wise pipeline, MultiCounter can extract more informative features, leading to better Period-mAP and counting accuracy.
\begin{table}[]
\caption{\textbf{Performance and convergence speed comparison.}}
\renewcommand\arraystretch{1.2}
\centering
\setlength{\tabcolsep}{1.2mm}{
\begin{tabular}{lcccc}
\bottomrule
Method             & Epochs & Period-mAP & AvgMAE $\downarrow $ & AvgOBO $\uparrow $ \\
\hline
Frame-wise \cite{wang2021end}           &   150  & 8.57      & 0.37     & 0.42     \\
\textbf{AoI-wise (Ours)} & 12  & \textbf{10.60}        & \textbf{0.24}     & \textbf{0.55}     \\
\toprule
\end{tabular}}
\label{tab:cs}
\end{table}

\textbf{Counting for Period-varying Actions.} We also evaluate the robustness of counting modules of MultiCounter (\emph{i.e.,} the task-specific heads). To facilitate the experiment, we have increased the proportion of video samples containing actions with varying periods in our MultiRep dataset. As shown in Table \ref{tab:tc}, compared to the combination between our MSTI module with existing single-repetition counters (\emph{e.g.,} RepNet \cite{dwibedi2020counting} and TransRAC \cite{hu2022transrac}), MultiCounter is not limited by periodic dynamics due to the tracking and counting joint optimization, resulting in better Period-mAP and counting accuracy.
\begin{table}[]
\caption{\textbf{Performance comparison for period-varying actions.}}
\renewcommand\arraystretch{1.2}
\centering
\setlength{\tabcolsep}{0.6mm}{
\begin{tabular}{lccc}
\bottomrule
Method             & Period-mAP & AvgMAE $\downarrow $ & AvgOBO $\uparrow $ \\
\hline
MSTI+RepNet \cite{dwibedi2020counting}           &   8.36      & 0.38     & 0.42     \\
MSTI+TransRAC \cite{hu2022transrac}           &   4.85      & 0.36     & 0.39     \\
\hline
\textbf{MSTI+Task-specific Heads (Ours)}  & \textbf{10.60}        & \textbf{0.24}     & \textbf{0.55}     \\
\toprule
\end{tabular}}
\label{tab:tc}
\end{table}
\begin{figure}[htbp]
		\hspace*{0cm}
 		\centering
 		\includegraphics[scale=0.57]{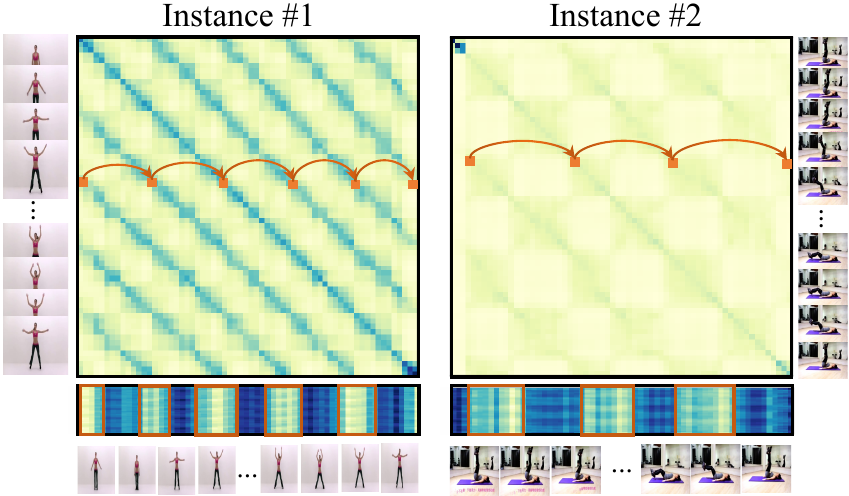}
 		\caption{\label{Fig2} \textbf{Visualizations of CDM output.} The square matrix depicts feature similarity scores ($S$) among frames. The rectangular color band shows attention scores ($A$) over time. Yellow markers indicate the periodic segments of each repetition.}
		\label{fig:att}
\end{figure}
\begin{figure*}[bhtp]
    \hspace*{0cm}
    \centering
    \includegraphics[scale=1.555]{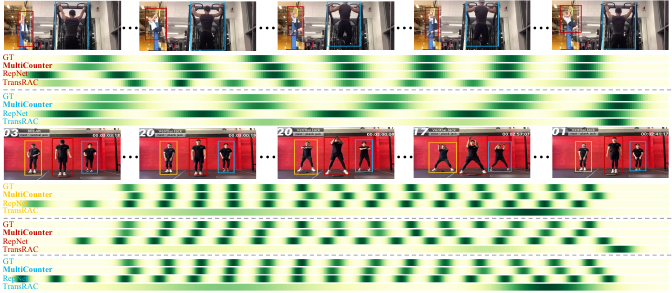}
    \caption{\label{Fig2}
\textbf{Qualitative examples of the instance-level repetition proposals predictions.} \textbf{Top:} Video from the synthetic MultiRep dataset, \textbf{Bottom:} Real video from in the wild (\emph{i.e.,} YouTube). For the results of SRAC methods, ByteTrack \cite{zhang2022bytetrack} is also utilized for action tracklet production.}
     \label{fig:vis}
\end{figure*}

\subsection{Visualizations}
As the intermediate representation bottleneck, CDM module helps model generalization to unseen activities during training. Figure \ref{fig:att} provides visualization of $S$ and $A$ in CDM. Firstly, CDM makes period velocity and periodicity predictions temporally interpretable so that easily derives repetition counts (\emph{e.g.,} Instance \#1 performs five repetitions and Instance \#2 does three times). Secondly, the mutual information captured in $S$ and $A$ allows compensation when one component has weaker signals. We also provide several examples of the instance-level repetition proposals in Figure \ref{fig:vis}. It shows that MultiCounter precisely delineates repetition proposals for instance-level actions while other methods struggle, confirming its generalization in real-world scenarios.

\subsection{Applications}
To explore MultiCounter's potential for downstream applications, we further experiment with the evaluation in the following aspects.

\begin{figure}[htbp]
		\hspace*{0cm}
 		\centering
 		\includegraphics[scale=0.5]{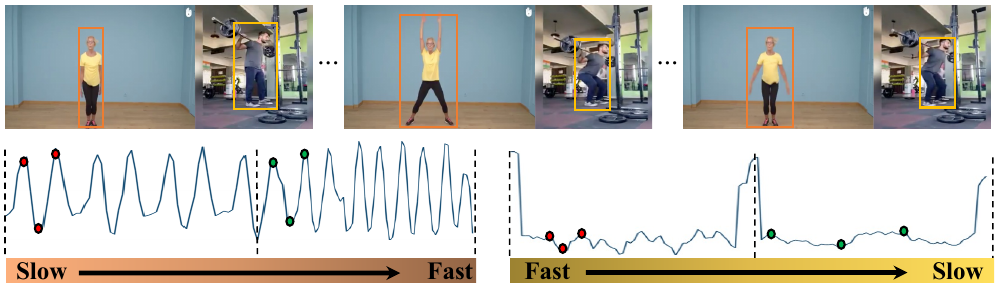}
 		\caption{\label{Fig2} \textbf{Estimation of speed rates of repetitions.} By projecting the 1D PCA features over time, MultiCounter enables accurate perception of speed changes in different instances.}
		\label{fig:pe}
\end{figure}

\textbf{Speed Rates Estimation of Repetitions \cite{dwibedi2020counting}}. The estimation of speed rates can reveal the continuous variations in movement frequency of multiple repetitions over time. As shown in Figure \ref{fig:pe}, by taking embeddings as 1D PCA projections, MultiCounter can infer whether an individual is speeding up or slowing down. This assists in monitoring speed changes of multiple simultaneous repetitive actions in physical exercise.

\begin{figure}[htbp]
		\hspace*{0cm}
 		\centering
 		\includegraphics[scale=0.53]{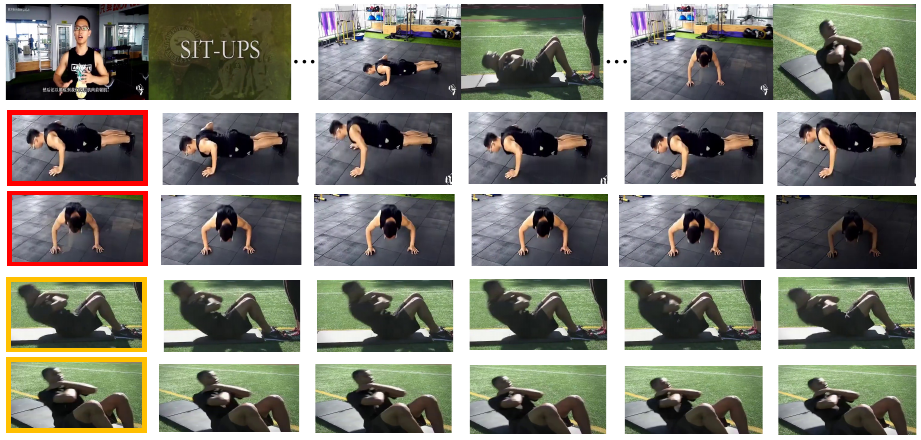}
 		\caption{\label{Fig2}\textbf{Qualitative results of cross-period multi-instance retrieval.} Given different instance queries performing repetitive actions (\emph{i.e., the red and yellow rectangular boxes}), the learned embeddings of MultiCounter can find similarities across different periods.}
		\label{fig:csr}
\end{figure}

\textbf{Cross-period Multi-instance Retrieval \cite{li2017measuring}}. The features extracted by MultiCounter can also facilitate cross-period multi-instance retrieval. By the K-nearest neighbor distances matching (\emph{K=1}), Figure \ref{fig:csr} shows that MultiCounter can retrieve the objectives similar to specific actions across different periods when given multiple action queries, suggesting its potential in downstream applications such as image retrieval.

\section{Conclusion}
In this paper, we formally define and explore the task of repetitive action counting for multiple instances in untrimmed videos, setting a new benchmark in the field of MRAC. To this end, we propose MultiCounter, a fully end-to-end deep learning framework tailored to address the fundamental challenges of MRAC, which not only simultaneously detects and tracks multiple human instances but robustly counts the number of repetitive actions. By training MultiCounter on the MultiRep dataset, our model’s superiority has been demonstrated substantially. Compared to ByteTrack+RepNet, a solution that combines an advanced tracker with a single repetition counter, MultiCounter substantially improves model performance. Additionally, MultiCounter runs in real-time on a commodity GPU server and is insensitive to human numbers due to its nature of one-stage inference. However, MultiCounter is designed for multiple action-agnostic repetition counting of human instances. In the future, we will focus on covering repeating signals of more object categories.

\begin{ack}
This work was supported in part by the National Key Research and Development Program of China under Grant 2022YFF0604504; in part by the National Science Foundation of China under Grant 62172439; in part by the Major Project of Natural Science Foundation of Hunan Province under Grant 2021JC0004; in part by the National Science Fund for Excellent Young Scholars of Hunan Province under Grant 2023JJ20076; in part by the Postdoctoral Fellowship Program of CPSF under Grant Number GZC20240832, and in part by the Central South University Innovation-Driven Research Programme under Grant 2023CXQD061. We also thank Debidatta Dwibedi for his help with this work.
\end{ack}

\bibliography{ecai}

\end{document}